%% file: main.tex
\documentclass[10pt,journal,compsoc]{IEEEtran}

\usepackage{amsmath,amsfonts}
\usepackage{array}
\usepackage[caption=false,font=normalsize,labelfont=sf,textfont=sf]{subfig}
\usepackage{textcomp}
\usepackage{stfloats}
\usepackage{url}
\usepackage{verbatim}
\usepackage{graphicx}

\usepackage{cite}
\usepackage{url}
\usepackage{times}
\usepackage[pagebackref,breaklinks,colorlinks]{hyperref}
\usepackage{microtype}
\usepackage{cleveref}
\usepackage{tikz}

\usepackage{multirow}
\usepackage{tablists}
\usepackage{xcolor, colortbl}
\usepackage{color}
\usepackage{threeparttable}
\usepackage{booktabs}
\usepackage{adjustbox}
\usepackage{tabularx}
\usepackage{hyphenat}
\usepackage{longtable}
\usepackage{rotating,siunitx}
\usepackage{etoolbox}
\usepackage{pifont} 
\usepackage{bm} 

\usepackage{subfig}
\usepackage{epsfig}
\usepackage{caption}
\usepackage[misc]{ifsym}
\usepackage{makecell}
\usepackage{tikz}
\usepackage[edges]{forest}
\usepackage{amssymb}
\usepackage{placeins}
\usepackage[normalem]{ulem}

\usepackage{algorithm}
\usepackage{etoolbox}
\usepackage{listings}
\usepackage{float}
\usepackage[noend]{algpseudocode}
\usepackage{etoolbox}
\makeatletter
\AfterEndEnvironment{algorithm}{\let\@algcomment\relax}
\AtEndEnvironment{algorithm}{\kern2pt\hrule\relax\vskip3pt\@algcomment}
\let\@algcomment\relax
\newcommand\algcomment[1]{\def\@algcomment{\footnotesize#1}}
\renewcommand\fs@ruled{\def\@fs@cfont{\bfseries}\let\@fs@capt\floatc@ruled
  \def\@fs@pre{\hrule height.8pt depth0pt \kern2pt}%
  \def\@fs@post{}%
  \def\@fs@mid{\kern2pt\hrule\kern2pt}%
  \let\@fs@iftopcapt\iftrue}
\makeatother

\makeatletter
\patchcmd{\@algocf@start}
  {0.5\algomargin}
  {0pt}
  {}{}
\patchcmd{\@algocf@finish}
  {0.1\algomargin}
  {-10pt}
  {}{}
\makeatother

\def\eg{\emph{e.g. }}

\def\vs{\emph{vs. }}

\newcommand{\etal}{\textit{et al}.}

\definecolor{LightRed}{rgb}{1,0.92,0.92}
\definecolor{LightOrange}{rgb}{1,0.95,0.88}
\definecolor{LightYellow}{rgb}{1.0,1.0,0.84}
\definecolor{LightGreen}{rgb}{0.9,1.0,0.88}
\definecolor{LightCyan}{rgb}{0.9,1,1}
\definecolor{LightBlue}{rgb}{0.9,0.94,1}
\definecolor{LightIndigo}{rgb}{0.92,0.9,1}
\definecolor{LightMagenta}{rgb}{0.96,0.86,1}
\definecolor{DirtyWhite}{rgb}{0.96,0.96,0.96}
\definecolor{DetectionColor}{rgb}{0.30,0.44,0.74}
\definecolor{GroundingColor}{rgb}{0.96,0.76,0.26}
\definecolor{CaptionColor}{rgb}{0.86,0.51,0.27}
\newcommand{\RedCycle}{\protect\tikz[baseline=0.0ex]\protect\draw[red,fill=red] (0,0) -- (0.3em,0.6em) -- (0.6em,0) -- cycle;}
\newcommand{\RedPentagram}{\protect\tikz[baseline=0.0ex]\protect\tikz[baseline=-0.5ex]\protect\draw[red,fill=red,scale=0.13](90:1) -- (234:1) -- (18:1) -- (162:1) -- (306:1) -- cycle;}

\begin{document}

\title{OV-DINO: Unified Open-Vocabulary Detection with Language-Aware Selective Fusion}

\author{Hao~Wang,~Pengzhen~Ren,~Zequn~Jie,~Xiao~Dong,~Chengjian~Feng,~Yinlong~Qian,\\
Lin~Ma,~Dongmei~Jiang,~Yaowei~Wang,~Xiangyuan~Lan\IEEEauthorrefmark{1},~Xiaodan~Liang\IEEEauthorrefmark{1},~\IEEEmembership{Senior Member,~IEEE}

\IEEEcompsocitemizethanks{
    \IEEEcompsocthanksitem Hao Wang is with the School of Intelligent Systems Engineering, Shenzhen Campus of Sun Yat-sen University, Shenzhen 518107, China and Pengcheng Lab, Shenzhen 518000 (e-mail: wangh739@mail2.sysu.edu.cn, wanghao9610@gmail.com).
    \IEEEcompsocthanksitem Pengzhen Ren is with the School of Intelligent Systems Engineering, Shenzhen Campus of Sun Yat-sen University, Shenzhen 518107, China.
    \IEEEcompsocthanksitem  Xiao Dong is with the School of Artificial Intelligence, Zhuhai Campus, Sun Yat-Sen University, Zhuhai, P.R. China, 519082.
    \IEEEcompsocthanksitem Zequn Jie, ~Chengjian~Feng, Lin Ma, and Yinlong Qian are with Meituan Inc, China.
    \IEEEcompsocthanksitem Xiangyuan Lan, Dongmei Jiang, and Yaowei Wang are with Pengcheng Lab, Shenzhen 518000, China. Yaowei Wang is also with the School of Computer Science and Technology, Harbin Institute of Technology, Shenzhen, China.
    \IEEEcompsocthanksitem Xiaodan Liang is with the School of Intelligent Systems Engineering, Shenzhen Campus of Sun Yat-sen University, Shenzhen 518107, China, and Pengcheng Lab, Shenzhen 518000 (e-mail: liangxd9@mail.sysu.edu.cn).
    \IEEEcompsocthanksitem\IEEEauthorrefmark{1} Xiangyuan~Lan, and Xiaodan~Liang are the corresponding authors.
}}

\markboth{Journal of \LaTeX\ Class Files,~Vol.~xx, No.~xx, xx~xx}%
{Shell \MakeLowercase{\textit{et al.}}: Bare Demo of IEEEtran.cls for Computer Society Journals}

\IEEEtitleabstractindextext{
\input{texs/00_abstract}
}

\maketitle

\IEEEdisplaynontitleabstractindextext

\IEEEpeerreviewmaketitle

\input{texs/01_introduction}

\input{texs/02_related}

\input{texs/03_method}

\input{texs/04_experiment}

\input{texs/10_conclusion}

\bibliographystyle{IEEEtran}
\bibliography{main}

\end{document}

%% file: texs/00_abstract.tex
\begin{abstract}
Open-vocabulary detection is a challenging task due to the requirement of detecting objects based on class names, including those not encountered during training. Existing methods have shown strong zero-shot detection capabilities through pre-training and pseudo-labeling on diverse large-scale datasets. However, these approaches encounter two main challenges: (i) how to effectively eliminate data noise from pseudo-labeling, and (ii) how to efficiently leverage the language-aware capability for region-level cross-modality fusion and alignment. To address these challenges, we propose a novel unified open-vocabulary detection method called OV-DINO, which is pre-trained on diverse large-scale datasets with language-aware selective fusion in a unified framework. Specifically, we introduce a Unified Data Integration (UniDI) pipeline to enable end-to-end training and eliminate noise from pseudo-label generation by unifying different data sources into detection-centric data format. In addition, we propose a Language-Aware Selective Fusion (LASF) module to enhance the cross-modality alignment through a language-aware query selection and fusion process. We evaluate the performance of the proposed OV-DINO on popular open-vocabulary detection benchmarks, achieving state-of-the-art results with an AP of 50.6\% on the COCO benchmark and 40.1\% on the LVIS benchmark in a zero-shot manner, demonstrating its strong generalization ability. Furthermore, the fine-tuned OV-DINO on COCO achieves 58.4\% AP, outperforming many existing methods with the same backbone.
The code for OV-DINO is available at \href{https://github.com/wanghao9610/OV-DINO}{https://github.com/wanghao9610/OV-DINO}.

\end{abstract}


%% file: texs/01_introduction.tex
\section{Introduction}
\label{sec:introduction}
\input{figs/fig1_comparison}
\IEEEPARstart{T}{he} traditional object detection methods, such as Fast R-CNN \cite{girshick2015fast}, Faster R-CNN \cite{ren2015fasterrcnn}, Mask R-CNN \cite{he2017maskrcnn}, DETR \cite{carion2020detr}, and DINO \cite{zhang2022dino}, are typically trained on datasets with closed-set categories, This limits their ability to detect objects outside of the predefined categories, which is a significant constraint for real-world applications. To address this limitation, a new task known as Open-Vocabulary Detection (OVD) has been proposed, attracting significant attention from both academic and industrial communities. Open-vocabulary detection requires the ability to detect any object using class names, even including objects that have never been encountered during training. 
The development of OVD can be traced back to the introduction of Zero-Shot Detection (ZSD) by Bansal \etal \cite{bansal2018zsd}, where models are trained on a limited set of categories and evaluated on novel categories. Building upon ZSD, Zareian \etal \cite{zareian2021ovrcnn} further expanded the concept to OVD by leveraging a visual semantic space derived from image-text data, thereby enhancing the capability of category generalization. 
\input{figs/fig2_illustration}

Recent studies \cite{radford2021clip, jia2021align, yao2021filip} have catalyzed the development of open-world vision methodologies \cite{long2023capdet, xu2024mqdet, feng2022promptdet, liu2023gdino, li2022glip}, enabling the detection of objects outside pre-defined categories. They typically pre-train the model on large-scale detection and grounding datasets and then generate pseudo-labels for image-text data.  
This introduces two distinct challenges: 

\noindent (i) \textbf{Data noise from the pseudo-labeling on image-text data.} It is attributed to the limited vocabulary concept of the detection data, and models trained with such data have poor generalization ability, leading to inaccurate predictions in pseudo-labeling on the image-text data, as depicted in \Cref{fig:comparison}(a). The current methods such as GLIP\cite{li2022glip}, GLIPv2\cite{zhang2022glipv2}, and G-DINO\cite{liu2023gdino} treat detection as a grounding task. These methods enable pre-training the model on large-scale detection and grounding datasets, followed by the generation of pseudo-labels for image-text data.
However, the categories involved in these two types of datasets are still limited. The pseudo-labels generated by the pre-trained model inevitably introduce noise when dealing with novel classes in the image-text data. 

\noindent (ii) \textbf{Alignment between the object features and the category description.}
The OVD methods aim to detect corresponding objects based on a specific category description. The objects in images often exhibit diverse features, which poses a challenge in detecting/aligning them with the specific category description. For example, given the category description `a photo of cat', the model is expected to align the category description with cats of different breeds, sizes, colors, etc.
To tackle this challenge, GLIP \cite{li2022glip} introduces complex deep fusion to integrate visual features into textual features. G-DINO \cite{liu2023gdino} proposes a bidirectional cross-attention-based lightweight feature enhancer to improve the text embedding representation. These methods employ image features to dynamically enhance the text embedding for better modality alignment. However, when an image contains multiple objects of the same category, the visual features of these objects are confused in a single text embedding, making it difficult to align the text embedding with each object.
\input{tabs/method_comparsion}

To address both key challenges, we introduce a novel method called OV-DINO for open-vocabulary detection.
For the first challenge, we propose a Unified Data Integration (UniDI) pipeline to integrate diverse data sources into a unified detection-centric data format, and pre-train the model on large-scale datasets in an end-to-end manner, as depicted in \Cref{fig:comparison}(b).
To achieve this, we consider the image-sized bounding box as the annotation box for the image-text data. The detection nouns, grounding phrases, and image captions serve as categories for detection-centric unification.
By doing so, \emph{UniDI not only eliminates the requirement of pseudo-label generation on image-text data, but also enhances the vocabulary concept during the pre-training stage.}
For the second challenge, we propose a Language-Aware Selective Fusion (LASF) module for region-level cross-modality fusion and alignment. 
As shown in \Cref{fig:illustration}(b), the LASF module enhances the embedding representation by selecting the text-related object embeddings. It then injects the text-related object embeddings into queries to improve modality alignment. \emph{LASF allows the model to dynamically align the category description with diverse objects in images, leading to more accurate predictions.}
Moreover, we propose a simple adaptation of the supervised training procedure used in DINO \cite{zhang2022dino} to facilitate \textbf{one-stage end-to-end training} for open-vocabulary detection, requiring only minimal modifications to the existing framework. To verify the effectiveness, extensive experiments are conducted on the popular open-vocabulary detection datasets COCO \cite{lin2014coco} and LVIS \cite{AgrimGupta2019lvisds} under zero-shot and fine-tuning settings. The results demonstrate that OV-DINO achieves state-of-the-art performance on both datasets and settings.
To highlight the characteristics of our model, we compare OV-DINO with recent methods in terms of method type, modality fusion, and pseudo-label generation in \Cref{table:method_compare}. 

In summary, our main contributions are outlined as follows:
\begin{itemize}
    \item We present OV-DINO, a novel unified open vocabulary detection approach that offers superior performance and effectiveness for practical real-world application.
    \item We propose a Unified Data Integration pipeline that integrates diverse data sources for end-to-end pre-training, and a Language-Aware Selective Fusion module to improve the vision-language alignment of the model.
    \item The proposed OV-DINO shows significant performance improvement on COCO and LVIS benchmarks compared to previous methods, achieving relative improvements of +2.5\% AP on COCO and +12.7\% AP on LVIS compared to G-DINO in zero-shot evaluation. The pre-trained model and code will be open-sourced to support open-end vision development.
\end{itemize}

%% file: figs/fig1_comparison.tex
\begin{figure}[!ht]
    \centering
    \includegraphics[width=0.95\linewidth]{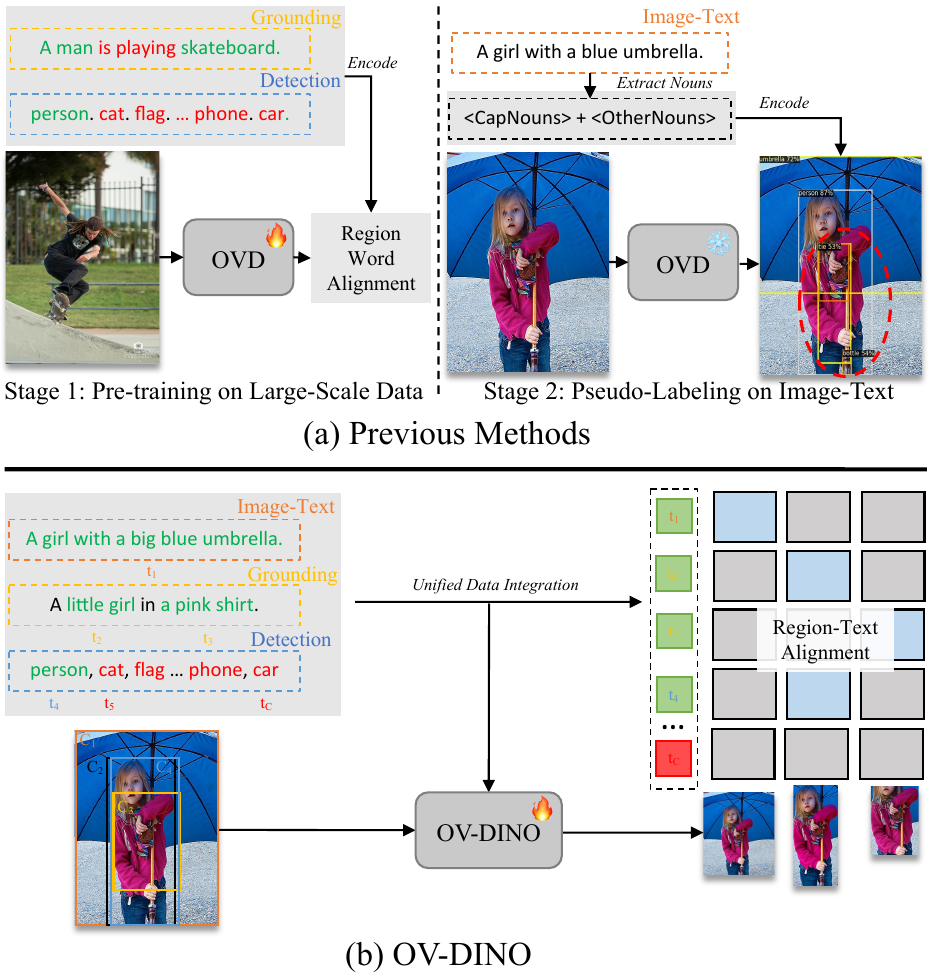}
    \caption{\textbf{Comparison of OV-DINO with Previous Methods.} (a) Previous methods (\eg GLIP \cite{li2022glip}, GLIPv2 \cite{zhang2022glipv2}, G-DINO \cite{liu2023gdino}) adopt a two-stage paradigm. They first pre-train on large-scale \textcolor{DetectionColor}{Detection} and \textcolor{GroundingColor}{Grounding} data, then generate pseudo labels on \textcolor{CaptionColor}{Image-Text} data, potentially introducing noise (\textcolor{red}{red} circle). (b) OV-DINO is a one-stage detection-centric method that integrates various data sources into a unified detection data format through a Unified Data Integration pipeline. It undergoes end-to-end pre-training via region-text alignment within a unified detection framework.}
    \label{fig:comparison}
    \vspace{-1.0cm}
\end{figure}

%% file: figs/fig2_illustration.tex
\begin{figure}[tb]
    \centering
    \includegraphics[width=0.98\linewidth]{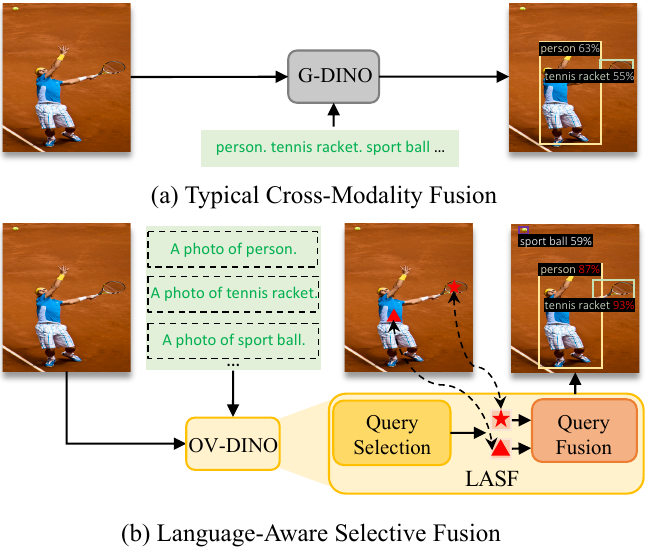}
    \caption{\textbf{Illustration of Language-Aware Selective Fusion (LASF).} We illustrate the processes of typical cross-modality fusion in G-DINO\cite{liu2023gdino} and language-aware selective fusion. LASF entails query selection and query fusion, which includes selecting the object embedding (\RedCycle\;,\RedPentagram\;) related to the text input, and fusing it with the learnable content query to improve prediction accuracy. In contrast, G-DINO directly fuses the query with text embedding. The OV-DINO with LASF achieves higher accuracy compared to G-IDNO (\eg \textcolor{red}{87\%} vs \textcolor{gray}{63\%} for ``person'', \textcolor{red}{93\%} vs \textcolor{gray}{55\%} for ``tennis racket''), highlighting the effectiveness of LASF in enhancing prediction accuracy.}
    \label{fig:illustration}
\end{figure}

%% file: tabs/method_comparsion.tex
\begin{table}[t]
    \centering
    \caption{\textbf{Comparison of OVD methods.} We compare OV-DINO with previous OVD methods in terms of method type, modality fusion, and pseudo-label generation. OV-DINO is a unified detection-centric method with LASF, eliminating the need for pseudo-label generation.}
    \begin{tabular}{l|c|c|c}
        \toprule
        Method & Type & Modality Fusion & Pseudo-Label \\ 
        \midrule
        GLIP\cite{li2022glip} & Grounding & DeepFusion & Y \\ 
        GLIPv2\cite{zhang2022glipv2} & Grounding & DeepFusion & Y \\ 
        G-DINO\cite{liu2023gdino} & Grounding & CrossAttnFusion & Y \\ 
        DetCLIP\cite{yao2022detclip} & Detection & -- & Y \\ 
        YOLO-World\cite{Cheng2024YOLOWorld} & Detection & RepVL-PAN & Y \\ 
        \midrule
        \rowcolor{LightBlue} OV-DINO(Ours) & Detection & LASF & N \\ 
        \bottomrule
    \end{tabular}
    \label{table:method_compare}
\end{table}

%% file: texs/02_related.tex
\section{Related Work}
\label{sec:related}
\textbf{Vision-Language Pre-Training.} Conventional supervised vision methods \cite{liu2021swin, ren2022beyond, he2016resnet, fu2019danet, wang2021tmanet} often rely on manual human annotation, thereby constraining the model's capacity for generalization.
Additionally, it is challenging to define a comprehensive list of categories and collect sufficient sample data for rare categories \cite{long2023capdet, ren2023viewco, wu2023baron}. The expensive labeling cost limits the wide application of the vision model for the open-world scenario.
To overcome the data annotation limitation, Vision-Language Pre-training has been proposed, it is a natural extension and development of the successful pre-train-then-fine-tune scheme in the domains of natural language processing (NLP) \cite{devlin2018bert, raffel2020t5} and computer vision \cite{dosovitskiy2020vit} community. Dual-stream approaches such as CLIP \cite{radford2021clip} and ALIGN \cite{jia2021align} have shown great zero-shot classification ability by pre-training on large-scale image-text pairs data (\eg CC12M \cite{sharma2018cc3m}, YFCC100M \cite{thomee2016yfcc100m}, Laion5B \cite{schuhmann2022laion}) with cross-modal contrastive learning. Single-stream approaches \cite{kim2021vilt, li2019visualbert, lu2019vilbert} directly model the relation of vision and text embedding by two separate transformer-based encoders, which perform well in tasks like image-text \cite{guo2020imgcap_0, yao2018imgcap_1, xu2015imgcap_2} and VQA \cite{antol2015vqa_0, gao2019vqa_1, li2020vqa_2}. Recently, VLMo \cite{bao2021vlmo}, BLIP \cite{li2022blip} and BLIPv2 \cite{li2023blip2} further explore a hybrid architecture incorporating both single-stream and two-stream architectures to facilitate a more cohesive way of vision-language understanding and generation. However, these models primarily focus on learning whole-image visual representations and cannot be directly applied to more complex core computer vision tasks such as segmentation and detection, which necessitate fine-grained semantic understanding. \\[-6pt]

\noindent \textbf{Open-Vocabulary Detection.} Traditional object detection methods \cite{girshick2015fast, ren2015fasterrcnn, he2017maskrcnn} have been successful in supervised scenarios, but face challenges in adapting to open-world scenarios with a large number of classes. It is challenging to explore approaches to acquire more semantic concepts for tasks related to Open-Vocabulary Detection (OVD). Recent approaches such as RegionCLIP \cite{zhong2022regionclip}, Baron \cite{zhong2022regionclip}, and ViLD \cite{gu2021vild} have concentrated on extracting intricate semantic correspondences and information to improve the inclusiveness of new categories. However, these approaches are based on the pre-trained CLIP model, which restricts their capacity for generalization. Furthermore, recent methods like GLIP \cite{li2022glip}, GDINO \cite{liu2023gdino}, and GLIPv2 \cite{zhang2022glipv2} aim to integrate multiple data sources to enrich the model's concept library. These approaches consider object detection as a grounding task and generate pseudo labels for image-text data. However, the grounding-orientated unification imposes limitations on the input length of text, and the pseudo-label generation introduces noise to the model. Meanwhile, DetCLIP \cite{yao2022detclip} proposes a dictionary-enriched visual-concept paralleled pre-training scheme to pre-train a model in a parallel way. DetCLIPv2 \cite{yao2023detclipv2} further endeavors to unify all data sources in a scalable pre-training approach by utilizing different losses for various data sources, sacrificing the efficiency of architecture. Therefore, this paper proposes a unified framework to integrate all data types into the object detection data format. The proposed approach aims to provide more accurate supervisory information to the model while overcoming text length limitations and the necessity for pseudo-label generation. This unified framework is designed to enhance the generalization of the model and improve the performance of open-vocabulary detection. \\[-6pt]

\noindent\textbf{Modality Information Fusion and Alignment.} Vision-Language model (VLM) has two distinct vision and language modalities, it is crucial to effectively fuse and align the modality information for VLMs. In the image-level VLMs, CLIP \cite{radford2021clip} and ALIGN \cite{jia2021align} directly align the vision and language modality with the contrastive loss \cite{hadsell2006contrative_loss}, FILIP \cite{yao2021filip} further aligns the modality information in fine-grained scale. To effectively align and fuse the cross-modal information, ALBEF \cite{li2021albef} proposes to align before fuse, which utilizes a multi-modal encoder to fuse the image features and text features through cross-modal attention and align the modality with an intermediate image-text contrastive loss. Flamingo \cite{alayrac2022flamingo} bridges the vision-only model and language-only model via the GATED XATTN-DENSE layers, achieving astonishing results on numerous benchmarks. For fine-grained cross-modal understanding, image-level modality fusion and alignment are insufficient for fine-grained vision-language understanding. In the region-level VLMs, RegionCLIP\cite{zhong2022regionclip} directly aligns the region representation with the region description via region-text pre-training, VLDet \cite{lin2022vldet} considers the region-text alignment as a bipartite matching problem. DetCLIP \cite{yao2022detclip} and DetCLIPv2 \cite{yao2023detclipv2} further extend the region-text alignment scheme via large-scale pre-training, achieving outstanding open-vocabulary detection performance. However, these approaches primarily concentrate on aligning modality information while ignoring the region-text modality fusion. To fuse the language information with the region representation, GLIP \cite{li2022glip} initially integrates cross-modal information in the encoder stage using a cross-attention module, then performs alignment using the region-word alignment loss. G-DINO \cite{liu2023gdino} further integrates modalities in the decoder stage. Although previous methods already consider fusion and alignment for cross-modal information interaction, they do not effectively balance the relationship between fusion and alignment. This paper aims to balance the fusion and alignment of modality information to enhance the model's ability to capture precise image details guided by language input.

%% file: texs/03_method.tex
\section{Method}
\input{figs/fig3_framework.tex}
This paper aims to develop a unified pre-training framework that integrates different data sources into a standardized format suitable for open-vocabulary detection tasks.
To accomplish this objective, we propose a novel model called OV-DINO, which leverages diverse data sources to improve the performance of open-vocabulary detectors within a unified pre-training framework (\Cref{method:overview}).
To facilitate unified pre-training across various data sources, we develop a Unified Data Integration (UniDI) pipeline applicable across various data sources (\Cref{method:unified_formulation}).
To fuse and align the fine-grained semantics between text embedding and region-specific visual embedding, we introduce a Language-Aware Selective Fusion (LASF) module to dynamically select and fuse the region-level vision-language information (\Cref{method:lasf}).
To enable detection-centric pre-training, we also develop a simple pre-training framework that features a straightforward design and shares similar training objectives with the closed-set detector DINO \cite{liu2023gdino} (\Cref{method:training}).

\subsection{Overview}
\label{method:overview}
The overall framework of OV-DINO is depicted in \Cref{fig:framework}, which includes a text encoder, an image encoder, and a detection decoder. Given an image accompanied by a prompt, the detection category nouns or grounding phrases are prompted to captions using specific templates to create a unified representation for the general text embedding. Subsequently, vanilla image and text embeddings are extracted using dedicated image and text encoders. After the embedding extraction, the vanilla image embedding along with the positional embedding are fed into the transformer encoder layers to generate refined image embedding. To improve the relevance between the image embedding and the text embedding, a language-aware query selection module is employed in detection decoder to select the object embedding associated with the text embedding. The selected object embedding serves as dynamic context embedding and is merged with static learnable content queries in the decoder using a language-aware query fusion module. The output queries from the final decoder layers are then used for classification projection and box regression to predict the corresponding classification scores and regress the object boxes. The model is pre-trained on diverse data sources (\eg detection, grounding and image-text data) to align the region-specific image embedding with the related text embedding in an end-to-end manner. It is optimized using the classification (alignment) loss and the box regression loss.

\input{figs/fig4_lasqf}
\subsection{Unified Data Integration}
\label{method:unified_formulation}
In the pre-training stage of OV-DINO, we leverage multiple data sources to enrich the semantic concept, encompassing detection, grounding, and image-text data. These data are annotated in different formats.
For example, the detection data is annotated with class labels and box coordination, the grounding data includes annotations of the caption with token positive indices and box coordination, and the image-text data solely consists of a text description for the image.
Typically, various type of data requires distinct processing methods, such as designing diverse loss functions for different data sources and generating pseudo-labels for image-text data.
This increases the complexity of model optimization, preventing the model from reaching its optimal performance. 
To tackle this problem, we propose a Unified Data Integration (UniDI) to convert all data sources into a unified detection-centric data format during data preparation process, thereby enabling the seamless integration of different types of data and harmonizing data from diverse sources for end-to-end training.
Integrating detection and grounding data is relatively straightforward, as grounding data can be considered a specific type of detection data, with each image having multiple grounding phrases.  
The challenge lies in seamlessly transforming large-scale image-text data into the detection data format. 
Drawing inspiration from Detic \cite{zhou2022detic}, we argue that the caption description of an image can be treated as a unique category for the image. Additionally, the annotation box for the image can be utilized as an image-sized bounding box. This innovative approach called \textbf{Caption Box}, enables the merging of these three types of data into a detection-centric data format.

To handle various data sources, we have established a standardized format for representing triplets of data as ($x$, $\{b_i\}_{i=1}^n$, $y$), where $x \in \mathbb{R}^{H \times W \times 3}$ represents the image input, $\{b_i \in \mathbb{R}^4 \}_{i=1}^n$ represents the bounding boxes, and $y \in \mathbb{R}^C$ represents the language text inputs. Here, $H$ stands for the image height, $W$ for the image width, and $n$ for the number of object instances. The bounding boxes $b_i$ are used as annotated boxes for detection and grounding data, while for image-text data, they represent the image-sized bounding box. The language text input $y$ varies depending on the data type. For detection data, it consists of pre-defined category names, for grounding data, it represents the grounding nouns or phrases of entities, and for image-text data, it is the entire caption. To ensure a consistent representation in language text embedding, we employ simple templates to prompt the text of detection and grounding data (\eg, a photo of \{category\}.), while leaving the text input for image-text data unchanged since it already serves as a caption. This approach is referred to as \textbf{Unified Prompt}, which enables all text inputs to be represented as the caption.

With the unified data integration pipeline (Caption Box and Unified Prompt), we can pre-train the model by combining training data from different data sources, including detection, grounding, and image-text data. Consequently, it eliminates the need for generating pseudo-labels on image-text data and enhances the vocabulary concept during the pre-training phase.

\subsection{Language-Aware Selective Fusion}
\label{method:lasf}
The open-vocabulary detection models aim to identify objects within an image by aligning the given text input with the semantic context of the image at the region level. 
However, objects in images often exhibit diverse semantic contexts, which presents a challenge in aligning the text input with these various semantic contexts. To overcome this challenge, we propose a Language-Aware Selective Fusion (LASF) module. 
This module dynamically selects the text-related object embeddings and injects them into the queries to improve modality alignment. The detailed architecture of LASF is depicted in \Cref{fig:lasf}(a). It comprises two essential components: language-aware query selection and language-aware query fusion.

The language-aware query selection component selects the object embedding by assessing the similarity between the image embedding and the text embedding. It computes the similarity of the multi-scale image embedding $E_{enc}$ and the text embedding $E_t$ and then choose the most relevant proposal embedding $E_{sp}$ and object embedding $E_{so}$. The selected proposal embedding is utilized to initialize the reference anchors, and the selected object embedding $E_{so}$ is forwarded for subsequent query fusion. 
The language-aware query selection can be formulated as follows:
\begin{align}
    E_{so}, E_{sp} &= RankTop(E_{enc} \otimes E_t^T),
\end{align}
where $E_t^T$ denotes the transpose of $E_t$, $\otimes$ denotes the Kronecker product\cite{van2000Kronecker}, and $RankTop$ is a parameter-less operation that arranges the elements in descending order and then selects the top $Q$ elements, $Q$ is the number of queries.

The language-aware query fusion component gradually fuses language-aware object embedding while preserving the original semantics of the content queries. This component is an essential part of the decoder layers and is repeated M times. Each decoder layer consists of several sub-layers including self-attention, cross-attention, gated-cross-attention, gated-feed-forward, and feed-forward layers. Initially, it takes the multi-scale image embedding $E_{enc}$, the selected object embedding $E_{so}$, and the learnable content query $Q_{lc}$ as input, and then dynamically updates the content query $Q_{lc}$. 
The language-aware query fusion can be formulated as follows:
\begin{align}
    Q^i_{lc_0} &= \Phi_{Attn}(qkv=Q^{i-1}_{lc}), \\
    Q^i_{lc_1} &= \Phi_{Attn}(q=Q^{i-1}_{lc_0}, kv=E_{enc}), \\
    Q^i_{lc_2} &= Q^i_{lc_1} + \tanh(\alpha_{a}) \ast \Phi_{Attn}(q=Q^i_{lc_1}, kv=E_{so}), \\
    Q^i_{lc_3} &= Q^i_{lc_2} + \tanh(\alpha_{b}) \ast \Phi_{FFW}(Q^i_{lc_2}), \\
    Q^i_{lc} &= \Phi_{FFW}(Q^i_{lc_3}),
\end{align}
where the upper index $i$ represents the module index, $\Phi_{Attn}$ represents the attention layer, $\Phi_{FFW}$ represents the feed-forward layer, $\alpha_{a}$ and the $\alpha_{b}$ are learnable parameters initialized to zero. This initialization ensures the training consistent with the original decoder framework while gradually incorporating language-aware context into the content query. 

To facilitate comprehension, we present the pseudocode for the Language-Aware Selective Fusion (LASF) in \Cref{alg:lasf}. 
We investigate three LASF variants depending on the placement of the object embedding: Later-LASF, Middle-LASF, and Early-LASF, as illustrated in \Cref{fig:lasf}. Additionally, the Typical Cross-Modality Fusion (Typical-CMF) proposed in G-DINO \cite{liu2023gdino} is also considered for comparison.

\input{algs/alg_lasf}
\subsection{Detection-Centric Pre-Training}
\label{method:training}
In this section, we present a one-stage end-to-end pre-training paradigm that integrates a variety of data sources.
Specifically, we utilize the proposed UniDI pipeline to convert diverse types of data into the detection-centric data format. This pipeline integrates data from multiple sources, including detection data, grounding data, and image-text data, facilitating the pre-training of a detection model with extensive semantic understanding. All the data sources adhere to a consistent model forward process and optimization losses, thereby achieving one-stage detection-centric pre-training in an end-to-end manner. \\[-6pt]

\noindent\textbf{Model Forward.} OV-DINO takes the triplet-wise data ($x$, $\{b_i\}_{i=1}^n$, $y$) as input. The image-encoder $\Phi_I$ is an image backbone to extract the image embedding $E_i \in \mathbb{R}^{P \times D}$ from the input image $x \in \mathbb{R}^{H \times W \times 3}$, where $P$ represents the spatial size of the flattened image embedding, $D$ represents the dimension of embedding. The text encoder $\Phi_T$ takes the language text $y \in \mathbb{R}^C$ as input and obtains the text embedding $E_t \in \mathbb{R}^{C \times D}$. The detection head of OV-DINO comprises a transformer encoder, a language-aware query selection module, and a transformer decoder with a language-aware query fusion module. The transformer encoder $\Phi_{Enc}$ takes encoded image embedding $E_i$ as input and outputs the refined multi-scale image embedding $E_{enc}$. The language-aware query selection module selects the most relevant image embedding according to the text embedding $E_t$ as the object embedding $E_{so} \in \mathbb{R} ^{Q \times D}$. The transformer decoder takes the learnable content query $Q_{lc} \in \mathbb{R} ^{Q \times D}$ as inputs, and interacts with the refined image embedding $E_{enc}$ and the selected object embedding $E_{so}$, which enables the query classification following the language text content.
After the decoder, a classification project layer $F_c$ projects the query embedding to a classification query logits $O \in \mathbb{R}^{Q \times D}$, and a regression layer $F_r$ predicts bounding boxes coordinates $B \in \mathbb{R}^{Q \times 4}$. Here, $Q$ and $C$ denote the length of queries and prompted captions, respectively. The classification alignment score matrix $S \in \mathbb{R}^{Q \times C}$ is obtained by calculating the similarity of $O$ and $E_t^T$. 
The overall process of model forward can be formulated as follows:
\begin{align}
    E_\mathrm{i} &= \Phi_\mathrm{I}(x), \; E_\mathrm{t} = \Phi_\mathrm{T}(y),\; E_\mathrm{enc} =\Phi_\mathrm{Enc}(\mathrm{E}_\mathrm{i}),\\
    E_\mathrm{so} &= \Phi_\mathrm{QS}(E_\mathrm{enc}, E_\mathrm{t}), \; Q_\mathrm{sf} = \Phi_\mathrm{QF}(E_\mathrm{enc}, E_\mathrm{so}, Q_\mathrm{lc}), \\
    O &= F_c(Q_\mathrm{sf}), \; B = F_r(Q_\mathrm{sf}), \; S = O \otimes E_\mathrm{t}^T,
\end{align}
where $E_t^T$ denotes the transpose of $E_t$, $\otimes$ means the Kronecker product\cite{van2000Kronecker}, $E_\mathrm{sp}$ is omitted for concise. \\[-6pt]

\noindent\textbf{Model Optimization.} The classification ground-truth $\mathrm{GT_{cls}} \in \{0, 1\}^{Q \times C}$ is a matrix that indicates the matched relationship between predicted regions and prompted texts. The bounding box ground-truth $\mathrm{GT_{box}} \in \mathbb{R}^{Q \times 4}$ is a matrix that contains corresponding box coordinates, they are constructed using the bipartite matching algorithm as described in \cite{zhang2022dino, carion2020detr}. The classification loss $\mathcal{L}_{cls}$ is calculated using the predicted alignment score $S$ and the ground-truth classification ground-truth $ \mathrm{GT_{cls}} $. The regression loss $\mathcal{L}_{reg}$ is calculated using the regressed bounding box $B$ and the bounding box ground-truth $\mathrm{GT_{box}}$. The regression loss encompasses both the box loss $\mathcal{L}_{box}$ and the generalized intersection over union (GIoU) loss $\mathcal{L}_{giou}$. In addition to the classification and regression losses, a denoising loss $\mathcal{L}_{dn}$ \cite{li2022dndetr} is introduced to enhance the stability of the training process. This loss function contributes to improving the robustness of the model during training. To maintain the simplicity of the detection-centric framework, the optimization objective of the pre-training stage is kept consistent with DINO \cite{zhang2022dino}. 
The whole optimization objective $\mathcal{L}$ is expressed as a combination of different loss components, and can be written as:
\begin{equation}
    \mathcal{L} = \alpha\mathcal{L}_{cls} + \beta\mathcal{L}_{box} + \gamma\mathcal{L}_{giou} + \mathcal{L}_{dn}. \\
\end{equation}
Here, $\alpha$, $\beta$ and $\gamma$ represent the weight factors of $\mathcal{L}_{cls}$, $\mathcal{L}_{box}$ and $\mathcal{L}_{giou}$, respectively. $\mathcal{L}_{cls}$ is implemented by a sigmoid focal loss\cite{lin2017focal}. $\mathcal{L}_{box}$ is implemented by an L1 loss. $\mathcal{L}_{giou}$ is implemented by a GIoU loss\cite{union2019giou}. $\mathcal{L}_{dn}$ represents the sum of the denoising losses \cite{li2022dndetr} of the label and box.

%% file: figs/fig3_framework.tex
\begin{figure*}[tp]
    \centering
    \includegraphics[width=\linewidth]{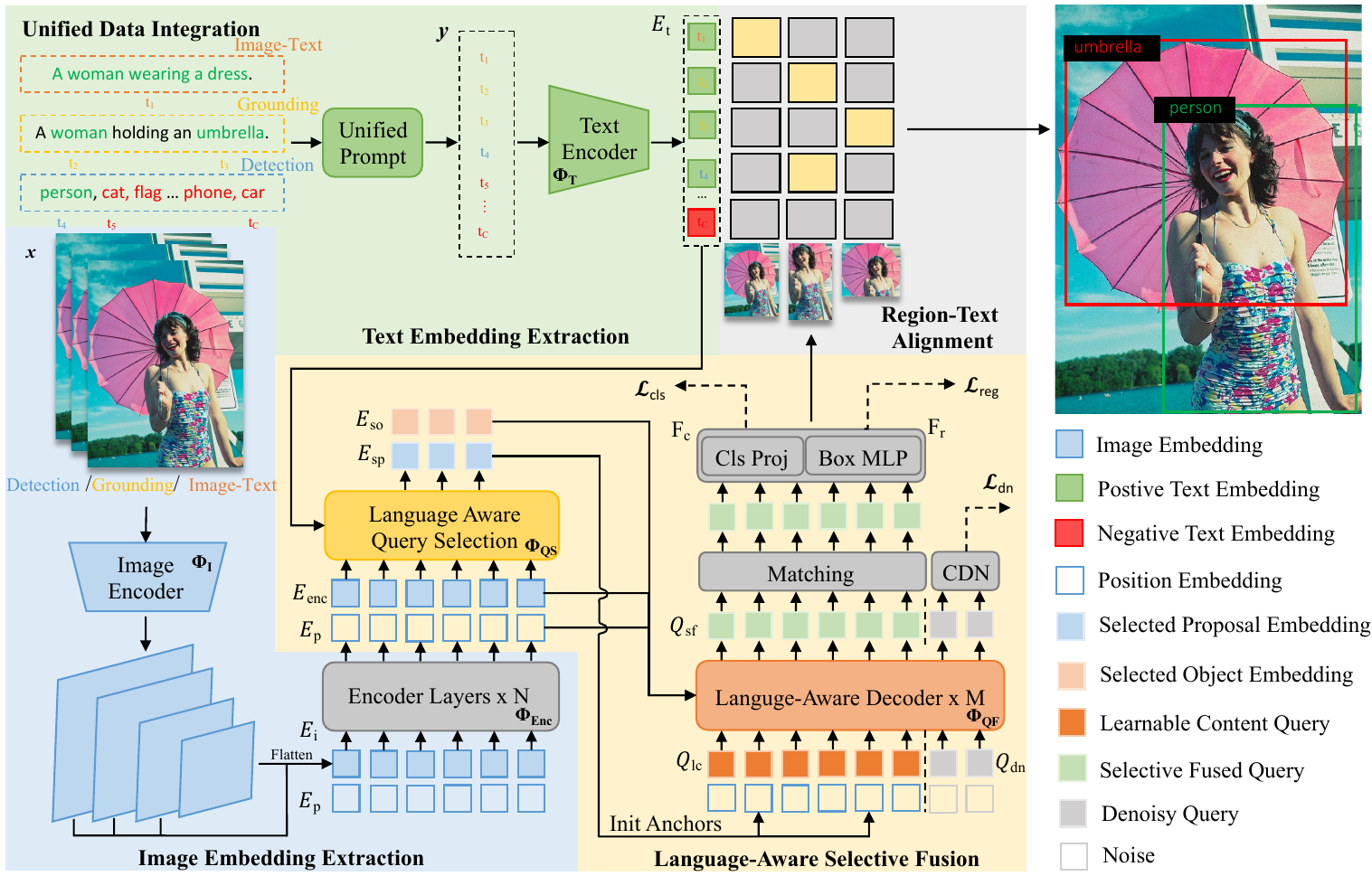}
    \caption{\textbf{Overall Framework of OV-DINO.} The pre-training of OV-DINO comprises three primary data sources (\textcolor{DetectionColor}{Detection}, \textcolor{GroundingColor}{Grounding}, \textcolor{CaptionColor}{Image-Text}). OV-DINO has three main components: a text encoder, a image encoder, and a language-aware detection decoder. First, we process the text inputs with \textit{Unified Data Integration} pipeline to ensure embedding representation consistency across these data sources. Then, the unified prompted text inputs go through a \textit{Text Encoder} to extract the text embedding, and the original image inputs undergo an \textit{Image Encoder} and some \textit{Encoder Layers} to output the multi-scale refined image embedding. Subsequently, we employ the \textit{Language-Aware Query Selection} to select the most relevant image embedding with the text embedding as the object embedding. The selected object embedding and the learnable content queries go through the \textit{Language-Aware Decoder} to fuse the content queries dynamically. Finally, OV-DINO outputs the classification scores by calculating the similarity of the projected query embedding with the text embedding through region-text alignment, and the regressed bounding boxes via an MLP layer.}
    \label{fig:framework}
\end{figure*}

%% file: figs/fig4_lasqf.tex
\begin{figure*}[t]
    \centering
    \includegraphics[width=\linewidth]{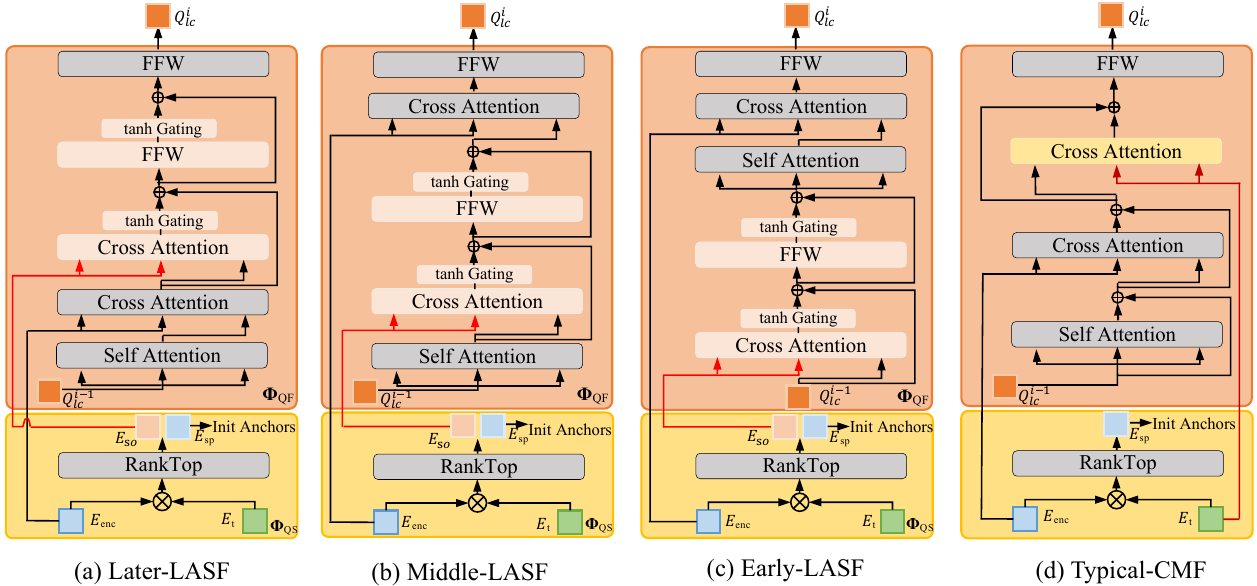}
    \caption{\textbf{Architecture of the Language-Aware Selective Fusion (LASF).} The LASF module consists of two main components: language-aware query selection $\bm{\Phi_{\text{QS}}}$ and language-aware query fusion $\bm{\Phi_{\text{QF}}}$. We illustrate three variants of the LASF module based on the insertion location of the object embedding: (a) Later-LASF, (b) Middle-LASF, and (c) Early-LASF. Additionally, we also illustrate (d) Typical-CMF proposed in G-DINO\cite{liu2023gdino} for clear comparison.}
    \label{fig:lasf}
\end{figure*}

%% file: algs/alg_lasf.tex
\begin{algorithm}[!tb]
\caption{Pseudocode of LASF in a PyTorch-like style.}
\label{alg:lasf}
\algcomment{\fontsize{8pt}{8pt}\selectfont \texttt{TopK}: topk selection; \texttt{Gather}: gathers values along index specified by dim; \texttt{Attn}: attention layer; \texttt{FFW}: feed-forward layer; \texttt{Tanh}: tanh activation function.}
\definecolor{codeblue}{rgb}{0.25,0.5,0.5}
\definecolor{codekw}{rgb}{0.85, 0.18, 0.50}
\lstset{
    backgroundcolor=\color{white},
    basicstyle=\fontsize{8pt}{8pt}\ttfamily\selectfont,
    columns=fullflexible,
    breaklines=true,
    commentstyle=\fontsize{8pt}{8pt}\color{codeblue},
    keywordstyle=\fontsize{8pt}{8pt},
}
\begin{lstlisting}[language=python]
def laqs(embed_enc, embed_t):
    """
    embed_enc: encoded embedding, shape: [B, P, D].
    embed_t: text embedding, shape: [B, C, D].
    """
    enc_box = BoxMLP(embed_enc)   #[B, P, 4]
    # RankTop operation.
    enc_cls = embed_enc @ embed_t.T #[B, P, C]
    topk_idx = TopK(enc_cls.max(-1)[0], Q, dim=1)
    # embed_so: [B, Q, D]
    embed_so = Gather(enc_cls, dim=1, topk_idx)
    # embed_sp: [B, Q, 4]
    embed_sp = Gather(enc_box, dim=1, topk_idx)

    return embed_so, embed_sp

def laqf(q_lc, embed_enc, embed_so):
    """
    q_lc: learnable content query, shape: [B, Q, D].
    embed_enc: encoded embedding, shape: [B, P, D].
    embed_so: object embedding, shape: [B, Q, D].
    """
    
    # self-attention
    q_lc = Attn(qkv=q_lc)
    # cross-attention
    q_lc = Attn(q=q_lc, kv=embed_enc)
    # gated-cross-attention
    q_lc = q_lc + Tanh(a) * Attn(q=q_lc, kv=embed_so)
    # gated-ffw
    q_lc = q_lc + Tanh(b) * FFW(q_lc)
    # ffw
    q_lc = FFW(q_lc)

    return q_lc

def lasf(embed_enc, embed_t, q_lc):
    """
    embed_enc: encoded embedding, shape: [B, P, D].
    embed_t: text embedding, shape: [B, C, D]. 
    q_lc: learnable content query, shape: [B, Q, D].
    
    NOTE: B is the batch size, P is the patch number, D is the dimension number, C is the prompted text number, and Q is the query number.
    """
    
    # 1. Language-aware query selection.
    embed_so, embed_sp = laqs(embed_enc, embed_t)
    # embed_sp to initialize the reference points, 
    # omit here for concise.

    # 2. Language-aware query fusion.
    # Decoder layers with laqf, iterate M times.
    for _ in range(M):
        q_lc = laqf(q_lc, embed_enc, embed_so)
    q_sf = q_lc
    
    return q_sf
    
\end{lstlisting}
\end{algorithm}

%% file: texs/04_experiment.tex
\section{Experiments}
\label{sec:experiments}
\input{tabs/pretrain_data}
In this section, we demonstrate the effectiveness of the proposed OV-DINO by conducting extensive experiments on two widely used open-vocabulary detection benchmarks: the COCO\cite{lin2014coco} and LVIS\cite{AgrimGupta2019lvisds}. We provide an overview of the pre-training datasets and the evaluation metrics in \Cref{exps:datasets}, and delve into the details of implementation in \Cref{exps:implementation}. We pre-train OV-DINO on large-scale diverse datasets and perform a zero-shot evaluation on the COCO and LVIS benchmarks. Following this, we fine-tune the pre-trained model on the COCO dataset and evaluate its performance in terms of close-set detection, as discussed in \Cref{exps:main_results}. To demonstrate the effectiveness of our model design, we conduct ablations in \Cref{exps:ablations}. Additionally, we present qualitative results for comparison with other methods, showcasing a clear representation of the detection results in \Cref{exps:visualizations}.

\subsection{Pre-Training Data and Evaluation Metric}
\label{exps:datasets}
\input{figs/fig5_capfilter}
\input{tabs/hyper_parameters}
\noindent\textbf{Pre-Training Data.}
In our experiments, we make use of several datasets as referenced in \cite{li2022glip, liu2023gdino, kamath2021mdetr}. These datasets comprise the Objects365 detection dataset \cite{shao2019objects365}, the GoldG grounding dataset \cite{kamath2021mdetr}, and the Conceptual Captions image-text dataset \cite{sharma2018cc3m}, as detailed in \Cref{tab:pretrain_data}. Our model is trained using the detection and grounding datasets following the methodology outlined in GLIP \cite{li2022glip}. However, the image-text dataset contains a significant amount of low-quality image-text pairs, as illustrated in \Cref{fig:capfilter}. The caption of the left sample effectively describes the image content, whereas the caption of the right sample does not align well with the image content. To mitigate the noise in the image-text dataset, we employ CLIP-Large \cite{radford2021clip} to filter 1 million image-text pairs from the original CC3M dataset. The filtering process begins by computing the similarity of 3 million pairs and subsequently ranking the top 1 million based on their image-text similarity. The effectiveness of the data filter is confirmed by the ablation study in \Cref{exps:ablations}. \\[-6pt]

\noindent\textbf{Evaluation Metric.} After pre-training, we evaluate the performance of the proposed OV-DINO under a zero-shot setting on the COCO \cite{lin2014coco} and LVIS \cite{AgrimGupta2019lvisds} benchmarks. In addition, we conduct further analysis by fine-tuning the pre-trained model on the COCO dataset to explore the effectiveness of continual fine-tuning. Following previous methods \cite{li2022glip, liu2023gdino}, we use the standard Average Precision (AP) metric to evaluate the performance of COCO, and the \textit{Fixed AP}\cite{dave2021fixedap} metric on LVIS for fair comparison.

\subsection{Implementation Details}
\label{exps:implementation}
\input{tabs/lvis_results}
\textbf{Model Architecture.} Constrained by the high cost of model training, we pre-train the model specifically using Swin-T \cite{liu2021swin} as the image encoder, which has shown superior performance compared to other methods. To ensure a fair comparison, we utilized the BERT-base from HuggingFace \cite{wolf2019huggingface} as the text encoder, consistent with the approaches used by GLIP \cite{li2022glip} and G-DINO \cite{liu2023gdino}. To incorporate category names in detection and noun phrases in grounding data during pre-training with image-text data, we adopted a unified data integration pipeline by prompting all category names or noun phrases with specific templates in CLIP \cite{radford2021clip}. Following DINO \cite{zhang2022dino}, we extracted multi-scale features at 4 scales ranging from 8x to 64x. Additionally, we set the maximum number of prompted text at 150, encompassing positive categories or phrases present in the image and randomly selected negative texts from all other data sources. For text embedding extraction, we employed the max-length padding mode and utilized mean pooling to aggregate text embedding along the length dimension. We integrated a linear projection layer to project the text embedding into the same embedding space as the query embedding. By default, we set the number of queries to 900, with six transformer layers in the encoder and decoder layers. \\[ -6pt ]

\noindent\textbf{Model Training.} To maintain simplicity in the model, we adhere to a similar training procedure as the original DINO setting \cite{zhang2022dino}. We adopt the AdamW \cite{loshchilov2017adamw} optimizer with a weight decay of 1e-4. The total batch size is 128, with a base learning rate of 2e-4 for all model parameters except the text encoder, which has a learning rate of 0.1 times the base learning rate (specifically set to 1e-5). During the fine-tuning stage on COCO, the base learning rate is adjusted to 1e-5, while the remaining hyper-parameters remain the same as in the pre-training stage. Both pre-training and fine-tuning are conducted for 24 epochs (2x schedule), using a step learning rate schedule where the learning rate is reduced to 0.1 and 0.01 of the base learning rate at the 16th and 22nd epochs, respectively. The weights allocated to the classification loss, box loss, and GIoU loss are 2.0, 5.0, and 2.0, respectively. The weights for matching cost components are identical to the losses except for the classification cost, which is given a weight of 1.0. The hyper-parameters used in the pre-training and fine-tuning stages of OV-DINO are detailed in \Cref{tab:hyper_parameters}.

\subsection{Main Results}
\label{exps:main_results}
\input{tabs/coco_results}
\input{tabs/ablation_modules}
\noindent\textbf{LVIS Benchmark.} In \Cref{tab:lvis_results}, we provide a comprehensive comparison of our proposed OV-DINO with recent state-of-the-art methods on the LVIS benchmark. The LVIS dataset is specifically designed to address long-tail objects and encompasses over 1000 categories for evaluation. Our evaluation of OV-DINO is conducted on the LVIS \texttt{MiniVal} and LVIS \texttt{Val} datasets under the zero-shot evaluation setting. OV-DINO surpasses previous state-of-the-art methods across various pre-training data settings. Specially, OV-DINO pre-trained on Objects365 (O365) dataset \cite{shao2019objects365} obtains superior results, with +5.9\% AP compared with GLIP. Combined with grounding data, OV-DINO demonstrates performance improvement, outperforming previous state-of-the-art methods, with +13.8\% AP and +5.0\% AP compared with G-DINO and DetCLIP. Moreover, when integrated with the image-text data, OV-DINO attains the highest AP results using the Swin-T image encoder under fair pre-training settings, setting a new record of 40.1\% AP on LVIS \texttt{MiniVal} and 32.9\% AP on LVIS \texttt{Val}. It is noteworthy that OV-DINO obtains +0.7\% AP gains using only image-text annotation, while other methods require pseudo-labeling for instance-level annotation. OV-DINO achieves superior performance with fewer parameters, showcasing its effectiveness and capability in detecting diverse categories. \\[-6pt]

\noindent\textbf{COCO Benchmark.} In \Cref{tab:coco_results}, we compare the proposed OV-DINO with recent state-of-the-art methods on the COCO benchmark in both zero-shot and fine-tuning settings. In the zero-shot setting, our models are pre-trained on various large-scale datasets and directly evaluated on the COCO dataset. Firstly, we pre-train the model on the O365 dataset and evaluate it using the zero-shot manner, where OV-DINO outperforms all previous models in the zero-shot evaluation setting, with +4.6\% AP and +2.8\% AP compared with GLIP and G-DINO, respectively. Remarkably, OV-DINO attains the best results when combined with the GoldG \cite{kamath2021mdetr} data, achieving 50.6\% AP in zero-shot transfer setting, outperforming YOLO-World +7.8\% AP and G-DINO +2.5\% AP. Additionally, we further fine-tune the pre-trained model on the COCO dataset, resulting in a new record of 58.4\% AP on COCO2017 validation set using only Swin-T \cite{liu2021swin} as the image encoder.  Significantly, OV-DINO undergoes pre-training for only 24 epochs, which is less than the pre-training schedules of other methods. Despite this, OV-DINO achieves the state-of-the-art performance in both zero-shot and fine-tuning settings. The outstanding performance achieved on COCO dataset illustrates that OV-DINO holds significant potential for practical applications. It's interesting to note that the addition of image-text data brings negative improvement to COCO, potentially due to the limited category names in the COCO dataset. Nevertheless, we find that image-text data is essential for discovering more diverse categories, as demonstrated in LVIS experiments.

\subsection{Ablation Study}
\label{exps:ablations}
\input{tabs/ablation_lasf}
\input{tabs/ablation_embedpool}
We conducted extensive ablation studies to analyze the effectiveness of the proposed OV-DINO. To reduce the cost of training with the full data, we randomly sampled 100,000 images from the original O365v1 \cite{shao2019objects365} dataset and 100,000 images from the filtered CC3M \cite{sharma2018cc3m} subset for all ablation studies. We set the batch size to 32 and the training schedule to 12 epochs. Unless specified, we pre-train OV-DINO on the sampled O365-100K and CC-100K datasets and evaluate zero-shot performance on the LVIS \texttt{MiniVal} dataset. \\[-6pt]

\noindent\textbf{Unified Data Integration.} In \Cref{tab:ablation_modules}, we conducte an ablation study on UniDI, which harmonizes different data sources through Unified Prompt and Caption Box. Unified Prompt utilizes specific templates to prompt category names, while Caption Box transforms image-text data into a detection-centric data format. The former results in +0.6\% AP gains (row 1 \vs row 0) for detection and +1.4\% AP gains (row 5 \vs row 4) for image-text data, and the latter led to +1.4\% AP gains (row 4 \vs row 2) by integrating image-text data. \\[-6pt]

\noindent\textbf{Language-Aware Selective Fusion.} In \Cref{tab:ablation_modules}, we also conduct an ablation study on LASF, which involves the dynamic selection and fusion of text-related object embeddings for region-level cross-modality fusion and alignment. LASF yields +0.9\% AP gains (row 2 \vs row 0), demonstrating the effectiveness of LASF. LASF, as a core module of OV-DINO, is able to continuously improve the performance on LVIS \texttt{MiniVal} together with UniDI. \\[-6pt]

\noindent\textbf{Variants of LASF.} In \Cref{tab:lasf}, we make a comparison of variants of the proposed LASF with the Typical-CMF in G-DINO\cite{liu2023gdino}. \Cref{fig:lasf} illustrates three variants of the LASF based on the insertion location of the object embedding: Later-LASF, Middle-LASF, and Early-LASF. Additionally, the architecture of Typical-CMF is provided for comparison. Extensive experiments are conducted to validate the effectiveness of LASF. All models in the ablations are pre-trained using a Swin-T as the image encoder on the sampled O365-100K subset. The results demonstrate that our LASF module is more effective in capturing language-aware context compared to the Typical-CMF module. Furthermore, the Later-LASF variant demonstrates superior zero-shot transfer ability on the LVIS \texttt{MiniVal} benchmark, which is adopted as our default architecture. \\[-6pt]

\noindent\textbf{Text Embedding Pooling.} In \Cref{tab:ablation_embedpool}, we evaluate the impact of different text embedding pooling methods, such as mean-pooling and max-pooling of the text embedding. The mean-pooling method computes the average value across the length dimension of the text embedding, while the max-pooling method identifies the maximum value along the token index in the text embedding. We pre-train the models on O365-100K and CC-100K with these two pooling methods, and it is observed that mean pooling demonstrates superior performance when applied to combined datasets. The mean-pooling method is effective in capturing the comprehensive representation of a prompted text, making it suitable for UniDI. \\[-6pt]

\noindent\textbf{Source of Image-Text Data.} In \Cref{tab:ablation_capsource}, we compare the performance across different sources of image-text data. We conducted the comparison by selecting the bottom and top 100K samples based on the image-text similarity of CLIP, as well as a random 100K sample. The results show that the rank\_top data source yields the best performance, while the rank\_bottom performs the worst. This highlights the inevitable noise in the image-text dataset and emphasizes the necessity of our filtering operation.

\input{tabs/ablation_capsource}
\subsection{Qualitative Results}
\label{exps:visualizations}
\noindent\textbf{Visualization on COCO.} We compare visualization results derived from the pre-trained OV-DINO with those from other methods. \Cref{fig:vis_coco} showcases the visualization results of zero-shot inference on the COCO dataset, where only the box predictions with a confidence score exceeding the threshold of 0.5 are displayed. Furthermore, a comparison is made with the predictions of GLIP \cite{li2022glip} and G-DINO \cite{liu2023gdino}. The first column depicts the image with ground truth, the second and third columns show the predictions of GLIP-T(B) and G-DINO-T$^3$, and the last column represents the predictions of OV-DINO$^2$, respectively. It is evident from the visualization that OV-DINO produces more precise predictions with higher confidence scores and is adept at detecting sufficient objects. These findings demonstrate the robust zero-shot transfer capability of OV-DINO in successfully detecting all objects based on the language text input. \\[-6pt]
\input{figs/fig6_coco_vis}
\input{figs/fig7_lvis_vis}

\noindent \textbf{Visualization on LVIS.} We also present visualization results derived from the pre-trained OV-DINO$^3$. \Cref{fig:vis_lvis} illustrates the visualization results of zero-shot inference on the LVIS dataset. The LVIS dataset is a long-tail dataset with more than 1000 categories, which can lead to numerous predictions in an image. For a clear visualization, we only display the box predictions with scores higher than 0.5. OV-DINO demonstrates exceptional performance in detecting a diverse range of categories, resulting in highly accurate predictions.

%% file: tabs/pretrain_data.tex
\begin{table}[tb]
    \centering
    \caption{\textbf{Pre-Training Data.} The dataset specifications used for pre-training OV-DINO. \# Texts denotes the number of categories for the detection dataset, the number of phrases for the grounding data, and the number of captions for the image-text dataset, respectively. \# Images denotes the number of images. \# Anno. denotes the number of instance annotations. CC1M$^\ddagger$ refers to our filtered 1M subset without any instance annotations.}
    \label{tab:pretrain_data}
    \small
    \begin{tabular}{llccc}
    \toprule
    Dataset & Type & \# Texts & \# Images & \# Anno. \\ 
    \midrule
    O365\cite{shao2019objects365} & Detection & 365 & 609K & 9621K \\ 
    GQA\cite{kamath2021mdetr} & Grounding & 387K & 621K & 3681K \\ 
    Flickr30k\cite{plummer2015flickr30k} & Grounding & 94K & 149K & 641K \\ 
    CC1M$^\ddagger$\cite{sharma2018cc3m} & Image-Text & 1M & 1M & -- \\ 
    \bottomrule
    \end{tabular}
\end{table}

%% file: figs/fig5_capfilter.tex
\begin{figure}[!tb]
    \centering
    \includegraphics[width=\linewidth]{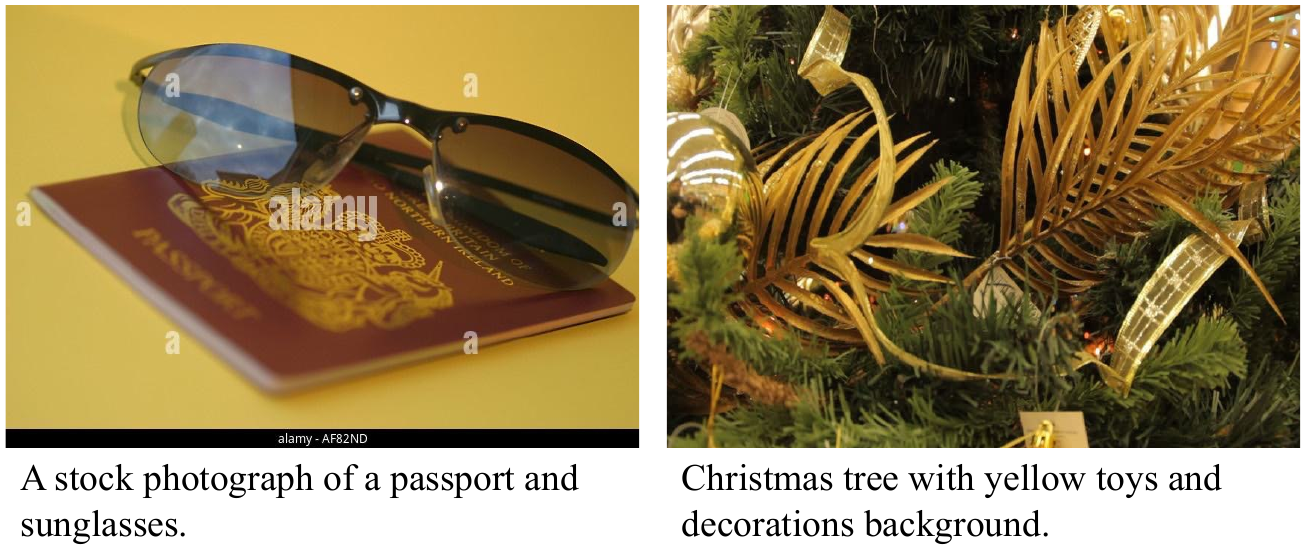}
    \caption{\textbf{Illustration of the Noise in the Image-caption Dataset.} The upper figure is the image, and the bottom text is the related caption for each sample. The sample on the left shows a high score of image-text similarity, while the sample on the right shows a lower score.}
    \label{fig:capfilter}
\end{figure}

%% file: tabs/hyper_parameters.tex
\begin{table}[!t]
    \centering
    \caption{\textbf{Hyper-Parameters in Pre-Training and Fine-Tuning of OV-DINO.} We emphasize the essential hyper-parameters for pre-training, while only addressing the distinct items of fine-tuning that differ from pre-training.}
    \label{tab:hyper_parameters}
    \small
    \begin{tabular}{l|r}
    \toprule
    Item & Value \\
    \midrule
    \multicolumn{2}{c}{Pre-Training Config} \\
    \midrule
    batch size & 128 \\ 
    training epochs & 24 \\ 
    optimizer & AdamW\cite{loshchilov2017adamw} \\ 
    weight decay & 1e-4 \\ 
    optimizer momentum & $\beta_1=0.9, \beta_2=0.999$ \\
    warmup iter & 1000 \\ 
    lr of image encoder & 2e-4 \\ 
    lr of text encoder & 2e-5 \\ 
    learning rate schedule & multi-step decay \\ 
    clip max norm & 0.1 \\ 
    input resolution & [800, 1333] \\ 
    hidden dim (D) & 256 \\
    \# of encoder layers (N) & 6 \\ 
    \# of decoder layers (M) & 6 \\ 
    \# of heads & 8 \\ 
    \# of queries (Q) & 900 \\ 
    \# of prompted text (C) & 150 \\
    cost of class & 1 \\ 
    cost of bbox & 5 \\ 
    cost of giou & 2 \\
    loss of class ($\alpha$) & 2 \\ 
    loss of bbox ($\beta$) & 5 \\ 
    loss of giou ($\gamma$) & 2 \\
    \midrule
    \multicolumn{2}{c}{Fine-Tuning Config} \\ 
    \midrule
    batch size & 32 \\ 
    lr of image encoder & 1e-5 \\ 
    lr of text encoder & 1e-6 \\ 
    \# of prompted text (C) & 80 \\
    \bottomrule
    \end{tabular}
\end{table}

%% file: tabs/lvis_results.tex
\begin{table*}[!htb]
    \centering
    \caption{\textbf{Zero-shot Domain Transfer Evaluation on LVIS \texttt{MiniVal} and \texttt{Val} Datasets(\%).} AP$_r$, AP$_c$, and AP$_f$ indicate the AP of rare, common and frequent categories, respectively. \textcolor{gray}{Gray} numbers denote that the model is trained on the LVIS dataset using either supervised or few-shot settings. CC3M$^\dagger$ denotes the pseudo-labeled CC3M in \cite{Cheng2024YOLOWorld}. CC1M$^\ddagger$ denotes a filtered subset from the CC3M dataset in our setting.}
    \label{tab:lvis_results}
    \small
    \begin{tabular}{l|c|c|c|cccc|cccc}
       \toprule
        \multirow{2}{*}{Model} & Image & \multirow{2}{*}{Params} & \multirow{2}{*}{Pre-Training Data} & \multicolumn{4}{c|}{LVIS MiniVal} & \multicolumn{4}{c}{LVIS Val} \\ 
        ~ & Encoder & ~ & ~ & AP & AP$_r$ & AP$_c$ & AP$_f$ & AP & AP$_r$ & AP$_c$ & AP$_f$ \\
        \midrule
        \color{gray} DETR\cite{carion2020detr} & RN101 & -- & \color{gray} LVIS & \color{gray} 17.8 & \color{gray} 3.2 & \color{gray} 12.9 & \color{gray} 24.8 & -- & -- & -- & -- \\
        \color{gray} MDETR\cite{kamath2021mdetr} & RN101 & \color{gray} 169M & \color{gray} GoldG, LVIS & \color{gray} 24.2 & \color{gray} 20.9 & \color{gray} 24.9 & \color{gray} 24.3 & -- & -- & -- & -- \\
        \color{gray} MaskRCNN\cite{he2017maskrcnn} & RN101 & -- & \color{gray} LVIS & \color{gray} 33.3 & \color{gray} 26.3 & \color{gray} 34.0 & \color{gray} 33.9 & -- & -- & -- & -- \\
        \midrule
        GLIP-T(A)\cite{li2022glip} & Swin-T & 232M & O365 & 18.5 & 14.2 & 13.9 & 23.4 & 12.3 & 6.0 & 8.0 & 19.4 \\
        GLIP-T(B)\cite{li2022glip} & Swin-T & 232M & O365 & 17.8 & 13.5 & 12.8 & 22.2 & 11.3 & 4.2 & 7.6 & 18.6 \\
        GLIP-T(C)\cite{li2022glip} & Swin-T & 232M & O365, GoldG & 24.9 & 17.7 & 19.5 & 31.0 & 16.5 & 7.5 & 11.6 & 26.1 \\
        GLIP-T\cite{li2022glip} & Swin-T & 232M & O365, GoldG, Cap4M & 26.0 & 20.8 & 21.4 & 31.0 & 17.2 & 10.1 & 12.5 & 25.5 \\
        G-DINO-T$^2$\cite{liu2023gdino} & Swin-T & 172M & O365, GoldG & 25.6 & 14.4 & 19.6 & 32.2 & -- & -- & -- & -- \\
        G-DINO-T$^3$\cite{liu2023gdino} & Swin-T & 172M & O365, GoldG, Cap4M & 27.4 & 20.8 & 21.4 & 31.0 & -- & -- & -- & -- \\
        DetCLIP-T(A)\cite{yao2022detclip} & Swin-T & 155M & O365 & 28.8 & 26.0 & 28.0 & 30.0 & 22.1 & 18.4 & 20.1 & 19.4 \\
        DetCLIP-T(B)\cite{yao2022detclip} & Swin-T & 155M & O365, GoldG & 34.4 & 26.9 & 33.9 & 36.3 & 27.2 & 21.9 & 25.5 & 31.5 \\
        DetCLIP-T\cite{yao2022detclip} & Swin-T & 155M & O365, GoldG, YFCC1M & 35.9 & 33.2 & 35.7 & 36.4 & 28.4 & 25.0 & 27.0 & 31.6 \\
        YOLO-World-S\cite{Cheng2024YOLOWorld} & YOLOv8-S & 77M & O365, GoldG & 26.2 & 19.1 & 23.6 & 29.8 & 24.2 & 16.4 & 21.7 & 27.8 \\
        YOLO-World-M\cite{Cheng2024YOLOWorld} & YOLOv8-M & 92M & O365, GoldG & 31.0 & 23.8 & 29.2 & 33.9 & -- & -- & -- & -- \\
        YOLO-World-L\cite{Cheng2024YOLOWorld} & YOLOv8-L & 110M & O365, GoldG, CC3M$^\dagger$ & 35.4 & 27.6 & 34.1 & 38.0 & -- & -- & -- & -- \\
        \midrule
        \rowcolor{LightBlue} OV-DINO$^1$(Ours) & Swin-T & 166M & O365 & 24.4 & 15.5 & 20.3 & 29.7 & 18.7 & 9.3 & 14.5 & 27.4 \\
        \rowcolor{LightBlue} OV-DINO$^2$(Ours) & Swin-T & 166M & O365, GoldG & 39.4 & 32.0 & 38.7 & 41.3 & 32.2 & 26.2 & 30.1 & 37.3 \\
        \rowcolor{LightBlue} OV-DINO$^3$(Ours) & Swin-T & 166M & O365, GoldG, CC1M$^\ddagger$ & \textbf{40.1} & \textbf{34.5} & \textbf{39.5} & \textbf{41.5} & \textbf{32.9} & \textbf{29.1} & \textbf{30.4} & \textbf{37.4} \\
        \bottomrule
    \end{tabular}
\end{table*}

%% file: tabs/coco_results.tex
\begin{table*}[!htb]
    \centering
    \caption{\textbf{Zero-shot Domain Transfer and Fine-tuning Evaluation on COCO(\%).} OV-DINO achieves superior performance than prior methods in zero-shot evaluation. Further fully fine-tuned on COCO, OV-DNIO surpasses the previous State-of-the-Art (SoTA) performance under the same setting. \textcolor{gray}{Gray} numbers denote the method is trained on the COCO dataset under the settings of supervised or few-shot.}
    \label{tab:coco_results}
    \small
    \begin{tabular}{l|c|c|c|c|c|c}
        \toprule
        \multirow{2}{*}{Model} & Image & \multirow{2}{*}{Pre-Training Data} &  \multirow{2}{*}{Data Size} & \multirow{2}{*}{Epochs} & \multicolumn{2}{c}{COCO 2017 Val} \\
        ~ & Encoder & ~ & ~ & ~ & Zero-Shot & Fine-Tuning \\
        \midrule
        \color{gray} Faster RCNN\cite{ren2015fasterrcnn} & \color{gray} RN50-FPN & \color{gray} COCO & \color{gray} 118K & \color{gray} 36 & -- & \color{gray} 40.3 \\
        \color{gray} Faster RCNN\cite{ren2015fasterrcnn} & \color{gray} RN101-FPN & \color{gray} COCO & \color{gray} 118K & \color{gray} 36 & -- & \color{gray} 41.8 \\
        \color{gray} DyHead-T\cite{dai2021dyhead} & \color{gray} Swin-T & \color{gray} COCO & \color{gray} 118K & \color{gray} 24 & -- & \color{gray} 49.7 \\
        \color{gray} DINO-T\cite{zhang2022dino} & \color{gray} Swin-T & \color{gray} COCO & \color{gray} 118K & \color{gray} 24 & -- & \color{gray} 51.3 \\
        \midrule
        GLIP-T(A)\cite{li2022glip} & Swin-T & O365 & 0.66M & 30 & 42.9 & 52.9 \\
        GLIP-T(B)\cite{li2022glip} & Swin-T & O365 & 0.66M & 30 & 44.9 & 53.8 \\
        GLIP-T(C)\cite{li2022glip} & Swin-T & O365, GoldG & 1.43M & 30 & 46.7 & 55.1 \\
        GLIP-T\cite{li2022glip} & Swin-T & O365, GoldG, Cap4M & 5.43M & 30 & 46.3 & 54.9 \\
        G-DINO-T$^1$\cite{liu2023gdino} & Swin-T & O365 & 0.61M & 50 & 46.7 & 56.9 \\
        G-DINO-T$^2$\cite{liu2023gdino} & Swin-T & O365, GoldG & 1.38M & 50 & 48.1 & 57.1 \\
        G-DINO-T$^3$\cite{liu2023gdino} & Swin-T & O365, GoldG, Cap4M & 5.38M & 50 & 48.4 & 57.2 \\
        YOLO-World-S\cite{Cheng2024YOLOWorld} & YOLOv8-S & O365, GoldG & 1.38M & 100 & 37.6 & 45.9 \\
        YOLO-World-M\cite{Cheng2024YOLOWorld} & YOLOv8-M & O365, GoldG & 1.38M & 100 & 42.8 & 51.2 \\
        YOLO-World-L\cite{Cheng2024YOLOWorld} & YOLOv8-L & O365, GoldG, CC3M$^\dagger$ & 1.63M & 100 & 45.1 & 53.3 \\
        \midrule
        \rowcolor{LightBlue} OV-DINO$^1$(Ours) & Swin-T & O365 & 0.60M & 24 & 49.5 & 57.5 \\
        \rowcolor{LightBlue} OV-DINO$^2$(Ours) & Swin-T & O365, GoldG & 1.38M & 24 & \textbf{50.6} & \textbf{58.4} \\
        \rowcolor{LightBlue} OV-DINO$^3$(Ours) & Swin-T & O365, GoldG, CC1M$^\ddagger$ & 2.38M & 24 & 50.2 & 58.2 \\
    \bottomrule
    \end{tabular}
\end{table*}

%% file: tabs/ablation_modules.tex
\begin{table*}[!htb]
    \centering
    \caption{\textbf{Ablations on Unified Data Integration and Language-Aware Query Fusion.} We evaluate the zero-shot performance on LVIS \texttt{MiniVal} of the proposed methods. UniDI, UniPro, and CapBox denote the Unified Data Integration, Unified Prompt, and Caption Box, respectively.}
    \label{tab:ablation_modules}
    \small
    \begin{tabular}{l|c|cc|c|cccc}
        \toprule
        \multirow{2}{*}{\#} & \multirow{2}{*}{Pre-Training Data} & \multicolumn{2}{c|}{UniDI} & \multirow{2}{*}{LASF} & \multirow{2}{*}{AP} & \multirow{2}{*}{AP$_r$} & \multirow{2}{*}{AP$_c$} & \multirow{2}{*}{AP$_f$} \\
        ~ & ~ & UniPro & CapBox & ~ & ~ & ~ & ~ & ~ \\
        \midrule
        0 & O365-100K & \textcolor{gray}{\ding{55}} & \textcolor{gray}{\ding{55}} & \textcolor{gray}{\ding{55}} & 18.3 & 10.1 & 14.8 & 22.8 \\
        1 & O365-100K & \ding{51} & \textcolor{gray}{\ding{55}} & \textcolor{gray}{\ding{55}} & 18.9 & 12.8 & 15.2 & 23.4 \\
        2 & O365-100K & \textcolor{gray}{\ding{55}} & \textcolor{gray}{\ding{55}} & \ding{51} & 19.2 & 10.5 & 16.5 & 23.1 \\
        3 & O365-100K & \ding{51} & \textcolor{gray}{\ding{55}} & \ding{51} & 19.5 & 12.8 & 16.6 & 23.4 \\
        \midrule
        4 & O365-100K, CC-100K & \textcolor{gray}{\ding{55}} & \ding{51} & \ding{51} & 20.6 & 13.1 & 17.9 & 24.4 \\
        5 & O365-100K, CC-100K & \ding{51} & \ding{51} & \ding{51} & \textbf{22.0} & \textbf{14.0} & \textbf{20.0} & \textbf{25.2} \\
        \bottomrule
    \end{tabular}
\end{table*}

%% file: tabs/ablation_lasf.tex
\begin{table}[tb]
    \centering
    \caption{\textbf{Ablations on Variants of Language-Aware Selective Fusion and Typical Cross-Modality Fusion.} We ablate the variants of LASF and Typical-CMF through the zero-shot LVIS \texttt{MiniVal} evaluation. All models are pre-trained on the O365-100K dataset.}
    \label{tab:lasf}
    \small
    \begin{tabular}{l|l|cccc}
        \toprule
        \# & Model & AP & AP$_r$ & AP$_c$ & AP$_f$\\
        \midrule
        0 & Baseline & 18.3 & 10.1 & 14.8 & 22.8 \\
        1 & Baseline + Typical-CMF & 18.9 & 10.4 & 16.0 & 22.9 \\
        2 & Baseline + Eearly-LASF & 18.8 & 9.5 & 16.1 & 22.9 \\
        3 & Baseline + Middle-LASF & 18.5 & 9.4 & 15.5 & 22.8 \\
        4 & Baseline + Later-LASF & \textbf{19.2} & \textbf{10.5} & \textbf{16.5} & \textbf{23.1} \\
        \bottomrule
    \end{tabular}
\end{table}

%% file: tabs/ablation_embedpool.tex
\begin{table}[tb]
    \centering
    \caption{\textbf{Ablations on Text Embedding Pooling.} We ablate the different text embedding pooling methods on O365-100K and CC-100K datasets, then evaluate zero-shot performance on LVIS \texttt{MiniVal}.}
    \label{tab:ablation_embedpool}
    \small
    \begin{tabular}{l|c|cc|cccc}
        \toprule
        \multirow{2}{*}{\#} & \multirow{2}{*}{Pre-Training} & \multicolumn{2}{c|}{EmbedPool} & \multirow{2}{*}{AP} & \multirow{2}{*}{AP$_r$}  & \multirow{2}{*}{AP$_c$} & \multirow{2}{*}{AP$_f$}\\
        ~ & ~ & mean & max & ~ & ~ & ~ & ~ \\
        \midrule
        0 & O365 & \textcolor{gray}{\ding{55}} & \ding{51} & 19.0 & 11.8 & 15.7 & 23.3 \\
        1 & O365 & \ding{51} & \textcolor{gray}{\ding{55}} & 18.9 & 10.7 & 15.1 & 23.7 \\
        \midrule
        2 & O365, CC & \textcolor{gray}{\ding{55}} & \ding{51} & 21.4 & 13.5 & 18.3 & \textbf{25.5} \\
        3 & O365, CC & \ding{51} & \textcolor{gray}{\ding{55}} & \textbf{22.0} & \textbf{14.0} & \textbf{20.0} & 25.2 \\
        \bottomrule
    \end{tabular}
\end{table}

%% file: tabs/ablation_capsource.tex
\begin{table}[tb]
    \centering
    \caption{\textbf{Ablations on the Source of Image-Text Data.} We ablate the different data sources of the image-text dataset and evaluate the zero-shot performance on LVIS \texttt{MiniVal}. The three data sources considered are: \textit{random\_select} entails randomly selecting 100K samples, \textit{rank\_bottom} and \textit{rank\_top} involve retaining the bottom 100K samples and the top 100K samples of the descending sorted image-text pairs, respectively.}
    \label{tab:ablation_capsource}
    \small
    \begin{tabular}{l|c|cccc}
        \toprule
        \# & Data Source & AP & AP$_r$ & AP$_c$ & AP$_f$ \\
        \midrule
        0 & rank\_bottom & 19.6 & 9.5 & 16.7 & 24.0 \\
        1 & random\_select & 20.8 & 11.6 & 18.1 & 24.8 \\
        2 & rank\_top  & \textbf{22.0} & \textbf{14.0} & \textbf{20.0} & \textbf{25.2} \\
        \bottomrule
    \end{tabular}
\end{table}

%% file: figs/fig6_coco_vis.tex
\begin{figure*}[!tp]
    \centering
    \includegraphics[width=\linewidth]{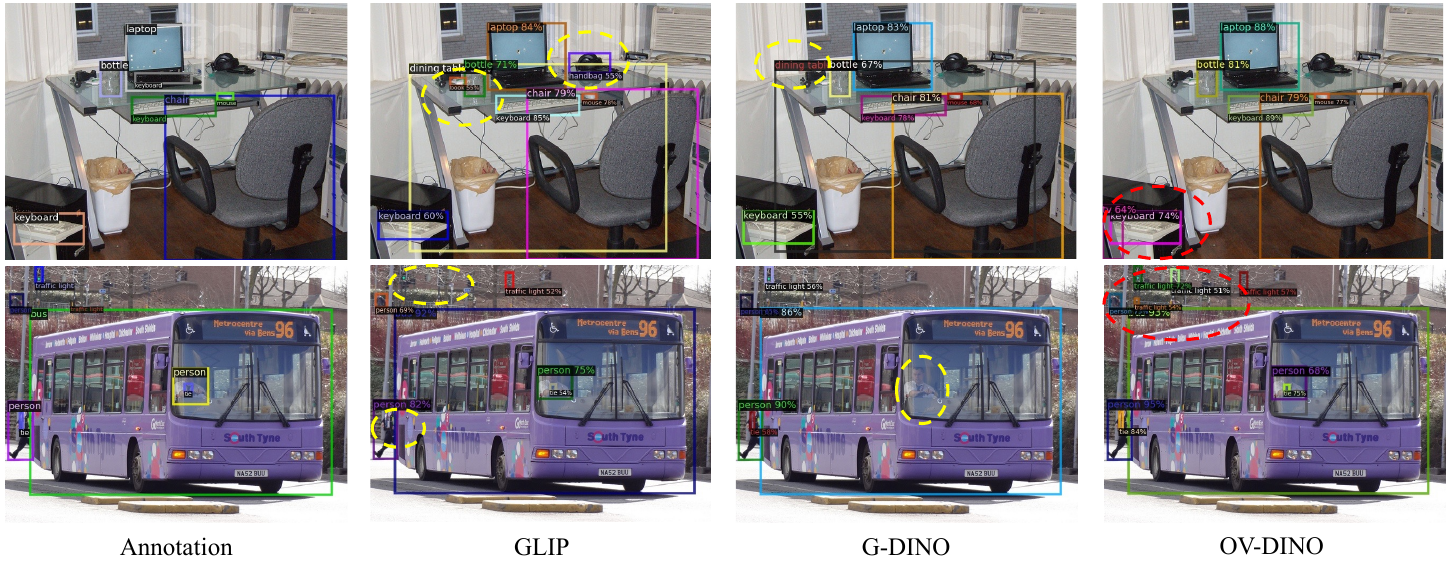}
    \caption{\textbf{Comparison of Visualization Results for Zero-Shot Inference on COCO.} We visualize the predictions of GLIP\cite{li2022glip}, G-DINO\cite{liu2023gdino} and the proposed OV-DINO. The failures are highlighted with a \textcolor{yellow}{yellow} circle. OV-DINO is capable of detecting all objects defined by COCO, and it can even detect additional objects that have not been labeled in the annotation (\textcolor{red}{red} circle).}
    \label{fig:vis_coco}
\end{figure*}

%% file: figs/fig7_lvis_vis.tex
\begin{figure*}[!htbp]
    \centering
    \includegraphics[width=\linewidth]{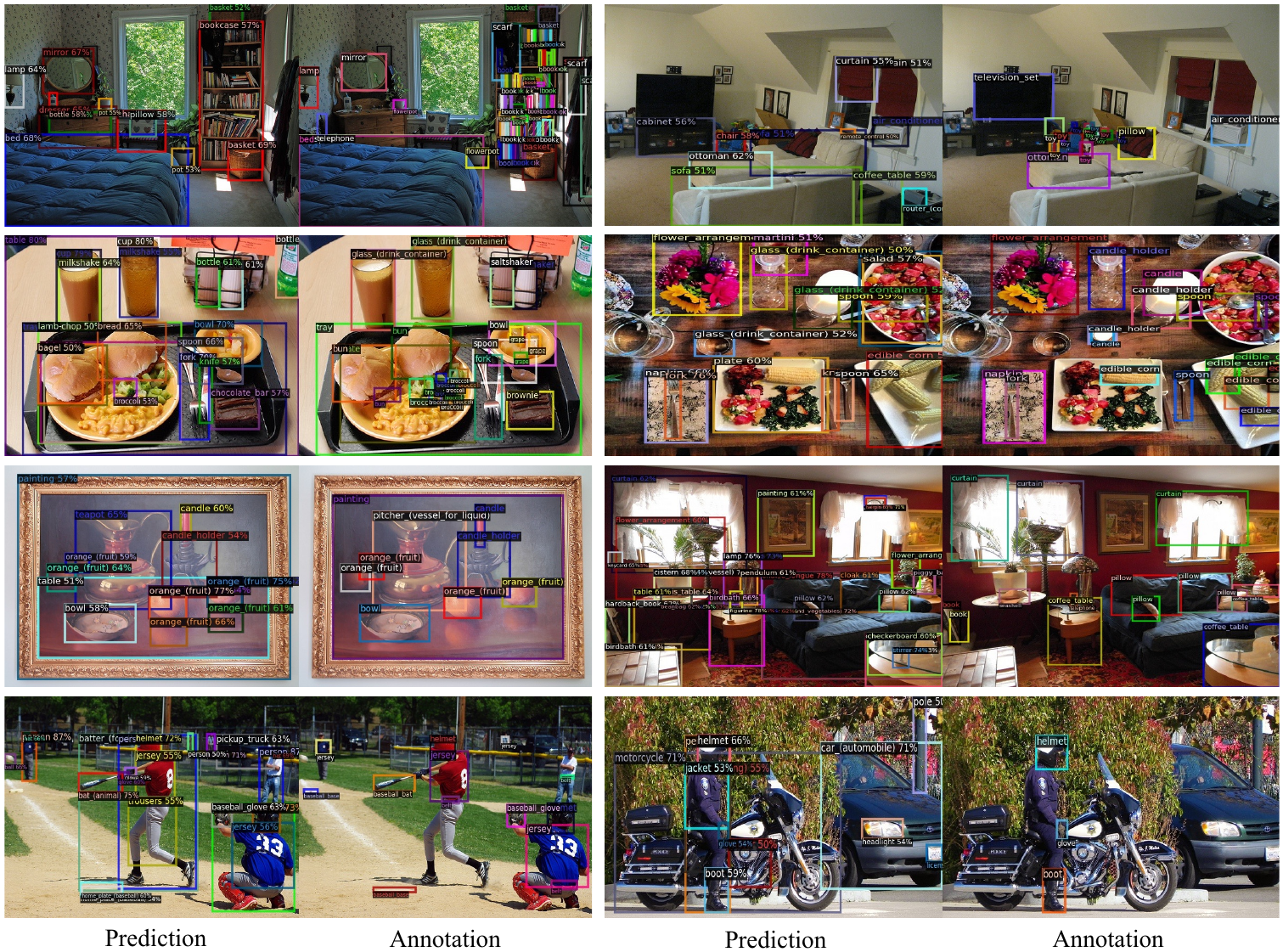}
    \caption{\textbf{Visualized Results for Zero-Shot Inference on the LVIS.} We visualize the predictions of OV-DINO, which shows a diverse range of instances being detected. Best viewed in zoom.}
    \label{fig:vis_lvis}
\end{figure*}

%% file: texs/10_conclusion.tex
\section{Discussions}
\noindent\textbf{Conclusions.} In this paper, we present OV-DINO, a robust unified open-vocabulary detector that aims to improve the performance of open-vocabulary detection. We propose a unified data integration pipeline to efficiently integrate various data sources, enabling end-to-end training with a unified detection framework for consistency and coherence. Additionally, we introduce a language-aware selective fusion module to selectively fuse cross-modality information, thereby improving the overall performance of OV-DINO. Experimental results demonstrate that OV-DINO outperforms previous state-of-the-art methods when evaluated on the challenging COCO and LVIS benchmarks. \\[-6pt]

\noindent\textbf{Limitations.} Despite the remarkable performance of OV-DINO as a unified open-vocabulary detection method, it is crucial to recognize that some specific challenges and limitations need to be addressed. One potential limitation is scaling up OV-DINO by incorporating a larger encoder and utilizing more extensive datasets. Scaling up shows a potential vision for improving the performance and applicability of the open-vocabulary detection model. However, it is inevitable to acknowledge that the pre-training stage requires substantial computational resources, which may present a barrier to scalability. Therefore, it is essential to strategically optimize the training process to facilitate the advancement of open-vocabulary tasks. \\[-6pt]

\noindent \textbf{Broader Impact.} In our research, we explore the detection-centric pre-training for open-vocabulary detection (OVD), which differs from the traditional approach of custom-designing for various data sources. Additionally, we introduce the concept of language-aware cross-modality fusion and alignment, marking a departure from the conventional method of simple region-concept alignment. Consequently, our research provides an innovative perspective for OVD. We expect that OV-DINO will encourage further exploration of ways to effectively leverage language-aware cross-modality information for open-vocabulary vision tasks.